\documentclass[10pt,twocolumn,letterpaper]{article}

 \usepackage{cvpr}              

\usepackage[accsupp]{axessibility} 
\usepackage{algorithm}
\usepackage{algorithmic}

\usepackage{subcaption}
\usepackage{multirow}
\usepackage{xcolor}
\usepackage{soul} 

\usepackage{amsmath}
\usepackage{amssymb}
\usepackage{mathtools}
\usepackage{amsthm}

\newcommand{\diff}[1]{{\color{black}{#1}}}

\definecolor{cvprblue}{rgb}{0.21,0.49,0.74}
\usepackage[pagebackref,breaklinks,colorlinks,citecolor=cvprblue]{hyperref}

\def\paperID{15132} 
\def\confName{CVPR}
\def\confYear{2024}

\title{Overload: Latency Attacks on Object Detection for Edge Devices}

\author{Erh-Chung Chen \textsuperscript{1} \thanks{The primary research and contribution for this work were conducted during a visit to IBM Research.}
\and
Pin-Yu Chen \textsuperscript{2}
\and
I-Hsin Chung \textsuperscript{2}
\and 
Che-Rung Lee \textsuperscript{1} 
\and
National Tsing Hua University \textsuperscript{1} \\
\and
IBM Research \textsuperscript{2} \\
}

\begin{document}
\maketitle

\begin{abstract}
  Nowadays, the deployment of deep learning-based applications is an essential task owing to the increasing demands on intelligent services. In this paper, we investigate \textit{latency attacks} on deep learning applications. Unlike common adversarial attacks for misclassification, the goal of latency attacks is to increase the inference time, which may stop applications from responding to the requests within a reasonable time. This kind of attack is ubiquitous for various applications, and we use object detection to demonstrate how such kind of attacks work. We also design a framework named \textit{Overload} to generate latency attacks at scale. Our method is based on a newly formulated optimization problem and a novel technique, called spatial attention. This attack serves to escalate the required computing costs during the inference time, consequently leading to an extended inference time for object detection. It presents a significant threat, especially to systems with limited computing resources. We conducted experiments using YOLOv5 models on Nvidia NX. Compared to existing methods, our method is simpler and more effective. The experimental results show that with latency attacks, the inference time of a single image can be increased ten times longer in reference to the normal setting. Moreover, our findings pose a potential new threat to all object detection tasks requiring non-maximum suppression (NMS), as our attack is NMS-agnostic.
\end{abstract}

\section{Introduction}
\label{sec:intro}
Deep neural networks (DNN) have accomplished many achievements in various fields \cite{he2016deep, devlin2018bert, brown2020language, bouman2016computational,van2020deep}. However, those models are usually large and demand a huge amount of computing resources, even for model inference. One solution is to utilize the computing power in cloud platforms, but the communication cost between the data center and the edge nodes can be high and the latency may not be acceptable for many real-time applications \cite{dai2019artificial,kong2022edge}.

In this paper, we investigate a new type of attacks, called latency attacks, whose purpose is to increase the execution time of the victim application. This kind of attacks poses critical threats to real-time applications, such as purchase behavior recognition system for unmanned store or autonomous cars systems, which are required to detect target objects and determine proper actions within stringent time constraints \cite{kim2020real,ravi2022real}. Any mistake or delayed response due to elongated latency could cause severe failures. Moreover, the latency attacks not only increase the execution time but also cause improper predictions made by the model indirectly.

The task we focus on is object detection, which has been widely used in numerous applications \cite{wang2021deep,dang2019deep, yang2021artificial}. Its purpose is to identify all objects in an image and to label them with corresponding classes and locations. Many deep learning-based models, such as SSD \cite{fu2017dssd} and YOLO series \cite{Jocher_YOLOv5_by_Ultralytics_2020}, often use Non-Maximum Suppression (NMS) as a post-processing operation to eliminate duplicated objects predicted by the detectors. The execution time of NMS depends on the number of objects fed into NMS. This property implies that object detection tasks may suffer from severe latency if numerous objects are predicted by the target model.

In this paper, we analyze the time complexity of NMS and find that the execution time is dominated by the total number of objects. The influence of the number of survival bounding boxes can be ignored. Based on the observation, we propose a Latency Attack on Object Detection (we name this attack framework ``Overload''), whose objective is to craft adversarial images with abundant objects from the victim model. To further improve the effectiveness of the adversarial image, we introduce a technique, called spatial attention, which can be used to manipulate the fake object generation in particular regions. This attack serves to escalate the required computing costs during the inference time, consequently leading to an extended inference time for object detection. It presents a significant threat, especially to systems with limited computing resources. 

Our main contributions are outlined as follows:
\begin{itemize}
    \item We systematically explore potential objectives for latency attacks, identifying the number of objects fed into NMS as a crucial factor for success, both theoretically and practically.
    \item Leveraging this insight, we simplify the objective design. In comparison to existing works like \cite{shapira2023phantom,wang2021daedalus}, Overload achieves superior attack performance while utilizing fewer computational costs and less memory.
    \item Experiments on Nvidia Jetson NX demonstrate that the inference time of our crafted images is approximately 10 times longer than that of the original images.
    \item We establish the NMS-agnostic nature of our proposed attack, underscoring its potential to pose a universal threat to all object detection systems reliant on NMS.
\end{itemize}

The rest of this paper is organized as follows. \Cref{sec:bkg} introduces the background on object detection and adversarial attack. \Cref{sec:method} presents the theoretical analysis of NMS and the proposed methods to craft the adversarial images. \Cref{sec:exp} shows the experimental results, and discussions on the impact of latency attacks for real-time applications. The last section is our conclusion. Analytical details of NMS and ablation studies are listed in the Appendix.

\begin{figure*}[t]
\centering
    \includegraphics[width=1.0\linewidth]{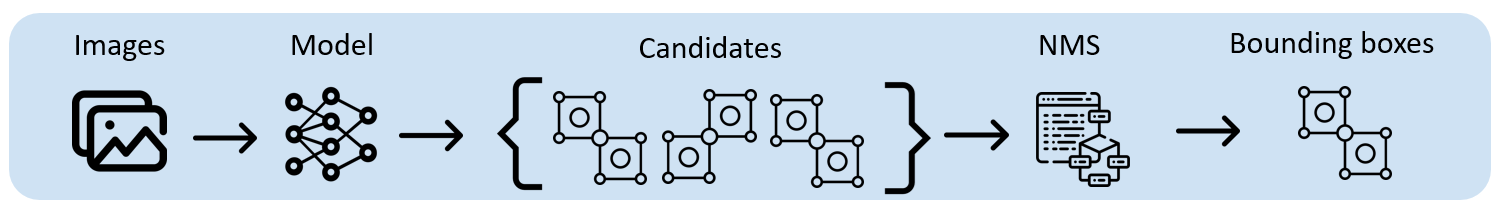}
    \vspace{-6mm}
    \caption{The processing flow of object detection. NMS stands for non-maximum suppression.}
    \label{fig:obj_pipeline}
\end{figure*}

\section{Background}
\label{sec:bkg}

\subsection{Object Detection}
Object detection is one of the challenging tasks in computer vision. In contrast to image classification, which receives an image and predicts its class, the goal of object detection is to recognize multiple objects and label their locations. To accomplish this, various deep learning-based detectors have been introduced. Some models approach this task by processing object locations and their probabilities in two distinct stages \cite{girshickICCV15fastrcnn,he2017mask}. Conversely, one-stage detectors tackle both classification and object localization within a single neural network \cite{Jocher_YOLOv5_by_Ultralytics_2020}.

To deploy those models onto edge devices, model size and inference time should be improved to meet the resource and timing constraints. Model compression is a commonly used strategy \cite{liu2021improving}. The YOLO family utilizes model scaling techniques to make the model fit different computing devices. Different scaling factors, such as resolution (size of the input image), depth (number of layers), width (number of channels), and stage (number of feature pyramids) are considered. For example, YOLO 5 has five different scales of models that use the same architecture but different width and depth configurations.

\subsection{Non-maximum Suppression}
\label{sec:bkg:nms}
Non-maximum suppression (NMS) is a prevalent component in modern object detection tasks \cite{girshickICCV15fastrcnn,Jocher_YOLOv5_by_Ultralytics_2020}. As shown in \Cref{fig:obj_pipeline}, an object detection model predicts numerous objects. Each object has its own information about height, width, position, objectness confidence, and probabilities of all classes. The objects with similar position information should be clustered as the same object. The main purpose of NMS is to eliminate redundant objects in each cluster and return survival objects as final bounding boxes. Several NMS implementations have been proposed, such as DIoU \cite{zheng2020distance} and CIoU \cite{zheng2020distance}.

While NMS is essential to most deep-learning-based models, there has been limited investigation into the elapsed time of NMS. This is primarily due to the fact that benchmarking datasets are typically clean and uncontaminated, and model performance is evaluated on high-end GPUs, where the time taken by NMS is negligible. Previous studies have mostly focused on optimizing performance through architecture refinements \cite{Jocher_YOLOv5_by_Ultralytics_2020}. However, when deploying models on edge devices, the computational cost of NMS becomes a critical consideration. Moreover, in this paper, we demonstrate that NMS can be exploited to launch latency attacks on object detection systems for edge devices.

\subsection{Adversarial Attack}
\label{sec:bkg:adv_atk}

Adversarial attacks have demonstrated that state-of-the-art models can be deceived by an adversarial example, which is the original input with an indistinguishable perturbation \cite{goodfellow2014explaining,carlini2018audio,gleave2019adversarial}. Adversarial attacks can be classified as white-box attacks \cite{Chen_2020_ACCV,chen2021ltd} or black-box attacks \cite{chen2017zoo,chen2023holistic}. In the white-box scenario, adversarial examples can be found along the gradient of a loss function. In the black-box scenario, the victim model is unaccessible. The gradient is often estimated by the finite differences method with multiple queries with stochastic perturbations.

Unlike the adversarial attack for image classification, whose goal is to degrade the model accuracy,  the objectives of adversarial attacks for object detection are diverse.  There are at least four objectives in current literature: appearing attack, hiding attack, mis-classifying attack, and mis-locating attack. Appearing attacks generate a perturbation such that the detector marks non-existent objects \cite{yin2022adc}. Hiding attacks make particular objects invisible \cite{wu2020making}. Mis-classifying attacks do not modify the information about the location but the predicted class \cite{cai2022context}. Mis-locating attacks focus on applications that require the locations of predicted bounding boxes to be highly precise. Tiny location changes can cause catastrophe failures \cite{jia2020fooling}.

\subsection{Latency Attack}
Latency attacks are a variant of adversarial attacks, with their primary objective being to maximize the processing time of the target model when presented with inputs. The Sponge attack, for instance, showcased the feasibility of latency attacks against deep learning-based translators operating on fixed-length sequences. In some dynamic vision models, predictions can be obtained from earlier stage if specific criteria are met \cite{kung2021add}. This behavior introduces a security concern regarding latency attacks \cite{hong2020panda}.

The vulnerability of NMS was addressed by Daedalus attack \cite{wang2021daedalus}, which aims to produce numerous bounding boxes and to reduce the mean average precision (mAP), a commonly used metric for object detection tasks. Another work is Phantom Sponge \cite{shapira2023phantom}, which demonstrates that increasing of the total number of objects fed into NMS can cause a longer execution time. It also tries to keep the recall of the modified image the same as that of the original image. Nonetheless, these approaches incur high computational costs due to the necessity of evaluating the gradients of all IoU pairs. Moreover, there is a lack of a comprehensive investigation into the impact of these attacks on edge devices.

\section{Overload: A Framework of Latency Attacks}
\label{sec:method}
\begin{figure}[t]
\centering
    \includegraphics[width=1.0\linewidth]{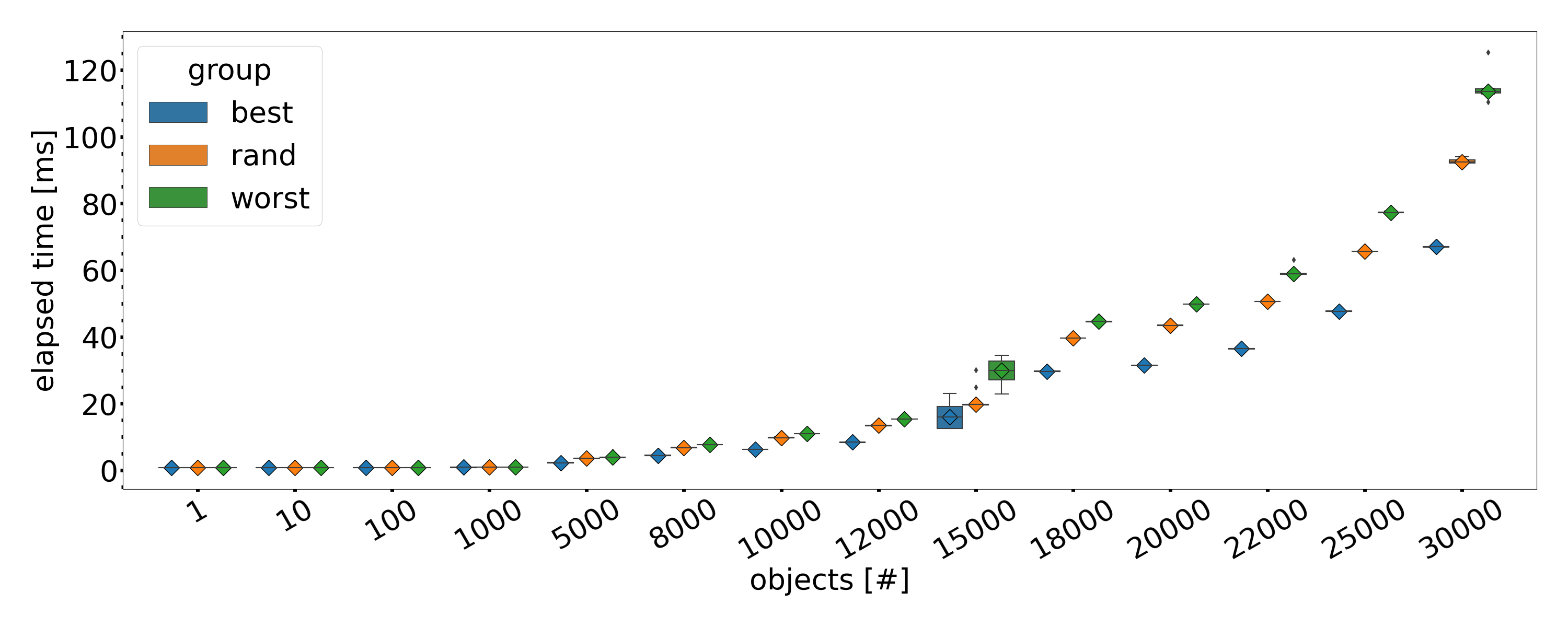}
    \caption{Elapsed time of NMS on NVIDIA Jetson NX.}
    \label{fig:obj_nms_nx}
\end{figure}
In this section, we first present the formal definition of latency attacks on object detection tasks and our motivations. We further provide theoretical explanations of how the adversarial examples are produced with minimal computational cost. Lastly, we propose an algorithm to generate adversarial images for latency attacks.

\subsection{Problem Definition and Motivation}
\diff{We assess potential vulnerabilities in object detectors within the white-box scenario, where the attackers have complete knowledge of the model architecture and can directly retrieve intermediate information from the victim model. Although this setting seems impractical in the real-world, it can serve as motivation to refine more robust model designs.}

The main goal of the latency attack is to find a perturbation joined to the input such that expected values of the elapsed time of every single request are maximized, which can be formulated as the following optimization problem:
\begin{equation}
\label{eq:goal}
     \mathop{\mathbb{E}}_{x \sim \mathcal{X}} \max_{\delta \leq \epsilon} T_{\textsubscript{total}}(x+\delta),
\end{equation}
where $\mathcal{X}$ is a set of requests to be processed and the perturbation $\delta$ is within the $\epsilon$-ball. Specifically, the elapsed time can be expressed as 
\begin{equation}
    T_{\textsubscript{total}}(x+\delta) = T_{\textsubscript{infer}}(x+\delta) + T_{\textsubscript{wait}}(\mathcal{X}),
\end{equation}
where $T_{\textsubscript{infer}}(x+\delta)$ is the inference time and $T_{\textsubscript{wait}}$ is the waiting time, which depends on the availability of the desired resources and all requests $\mathcal{X}$. We assume there is a queue for the requests. When the desired computing resources are fully occupied, requests will be stored in the queue. Therefore, the goal defined in \Cref{eq:goal} is equivalent to finding adversarial examples that could exhaust the desired resources so that the rest of the requests in the queue cannot be processed until the malicious requests are terminated. 

\diff{
However, directly incorporating the function $T$ into the formulation for generating adversarial attacks poses a challenge, because $T$ is a complex function that can be influenced by several external factors, such as scheduling or cache missing. In the case of white-box attacks, computing the derivative directly through the computational graph is not feasible. Moreover, for black-box attacks, discerning whether the increase in processing time stems from changes in machine status or the introduction of an adversarial example is non-trivial. In this paper, we seek to address the following two critical questions:
\begin{itemize}
    \item How can we parameterize the elapsed time and the outputs generated by the target model, allowing us to formulate an objective that can be exploited to launch latency attacks on object detection?
    \item How can we maximize the elapsed time of the victim model in processing the given image, all while minimizing computational costs and memory usage in the production of adversarial examples?
\end{itemize}
}

\subsection{Theoretical Analysis of NMS Time Complexity}
\label{sec:med:nms}

A fundamental requirement for a successful latency attack is the existence of an operation whose processing time is contingent on the input. In the case of object detection tasks, NMS emerges as the pivotal factor. To delve into the vulnerability of NMS in object detection to latency attacks, we first analyze the time complexity of NMS based on the commonly used deep learning frameworks on GPU, as suggested in \cite{torchvision2016}. Let $\mathcal{C}$ be the set of objects fed into NMS. The steps of NMS can be divided into four major tasks. We leave the detailed analysis in \Cref{sec:nms_algo} and summarize their time complexity below:.
\begin{enumerate}
  \item Filtering low-confidence objects: $O(|\mathcal{C}|)$;
  \item Sorting candidates by probabilities: $O(|\mathcal{C}| \log |\mathcal{C}|)$;
  \item Calculating pairwise IoU scores: $O(|\mathcal{C}|^2)$;
  \item Pruning inactive objects: $O(|\mathcal{C}|^2)$.
\end{enumerate}
As evident, the time complexity appears to be quadratic. However, in practice, the time is dominated by unavoidable overheads when $|\mathcal{C}|$ is relatively small . Hence, the elapsed time can be approximated as
\begin{equation}
\label{eq:comp}
    T_{\text{total}} \approx
    \left\{
    \begin{array}{ll}
    \alpha|\mathcal{C}|^2, & \mbox{if } |\mathcal{C}|> N\\
    T_{\text{base}}, & \mbox{otherwise},
    \end{array}
    \right.
\end{equation}
where $\alpha$ is the calibrated coefficient depending on box distributions,  $T_{\text{base}}$, and $N$ are constants related to the computational capacity of used GPU. 

To ascertain whether the time complexity is affected by the number of output boxes, we conducted experiments to measure the elapsed time of NMS using synthetic data under three distinct scenarios: worst, best, and random.  The worst case is simulated by setting $T_\text{IoU}$ to 1.0 so that no object can be filtered out after NMS. On the other hand, the best scenario is making all elements in set $\mathcal{C}$ share the identical box information, so that NMS can mark all objects in the first iteration.  The random scenario just randomly sets the box configurations.

\diff{
The experimental results are illustrated in \Cref{fig:obj_nms_nx}, and two key observations can be drawn from the findings. First, as anticipated, the processing time in all three cases displays a quadratic growth pattern as the size of $|\mathcal{C}|$ increases. Second, the impact of the number of output boxes can be represented by the ratio of the worst-case scenario to the best-case scenario, which remains a constant, regardless of the number of objects ($|\mathcal{C}|$). Therefore, we can formulate the elapsed time of NMS as
\begin{equation}
\label{eq:T_cost}
    T_{\text{total}} = s (1 + \frac{|\mathcal{B}|}{|\mathcal{C}|})|\mathcal{C}|^2 = s (|\mathcal{C}|^2 + |\mathcal{B}||\mathcal{C}|),
\end{equation}
where $s$ is the calibrated coefficient and $\mathcal{B}$ is the set of boxes output by NMS.
}

\subsection{Exploring Objectives for Adversarial Example Generation}
Our primary objective is to maximize the elapsed time of the victim model when processing a given image, all while minimizing computational costs and memory usage when creating adversarial examples. \Cref{eq:T_cost} implies that latency can be increased by generating more objects or boxes, but the associated cost of producing adversarial examples remains unclear.

\diff{
In pursuit of our goal, we aim to explore all potential objectives for generating adversarial examples. Previous works \cite{wang2021daedalus,shapira2023phantom} suggested that the latency attack can be achieved by a composite objective that involves three components: maximizing the confidence of objects, minimizing the pairwise IoU scores, and minimizing the areas of objects. The first terms can prevent the objects from being filtered at the first step of NMS. The second term drifts the centers of objects and the third term adjusts the areas of boxes to avoid the removal of objects in the pruning process.

However, the objective proposed in previous works overlooks the influence of the number of boxes. The observations in \Cref{fig:obj_nms_nx} indicate that the elapsed time can be increased by up to 20 times when generating numerous objects. On the other hand, the time increment is approximately 2 times when the number of boxes is increased with a fixed $|\mathcal{C}|$. 

Moreover, while the IoU computation is differentiable, tracking all pairwise comparisons consumes considerable memory and time for crafting adversarial examples. Furthermore, the performance improvement cannot be easily quantified, as the execution order influences the final results. For example, a modified box can evade the discarding of objects but it may cause some nearby boxes to be marked duplicated in the rest pruning rounds.
}

This insight reinforces our emphasis on prioritizing maximum object confidence as crucial for optimizing latency while minimizing computational costs and memory usage. We contend that including the last two terms in the objective not only fails to improve performance but also increases the time and memory required for generating adversarial examples, adding unnecessary complexity to the design.

\subsection{Proposed Method for Latency Attacks on Object Detection}
\label{sec:algo}

We propose that latency attacks on object detection can be formulated as an optimization problem, 
\begin{equation}
\label{eq:loss_fun}
    \max_x \sum^{n}_{i=0} F_l(F_{\text{conf}}(M(x)_i)),
\end{equation}
where $M(x)$ is the predicted results; $n$ is the total number of objects output by the given model, $F_l(\cdot)$ is a monotonic increasing function; and $F_{\text{conf}}(\cdot)$ retrieves each object's corresponding confidence. Specifically, 
\begin{equation}
\label{eq:conf}
    F_{\text{conf}}(\cdot) = 
    \begin{cases}
        c  ,& \text{if } cp_i > T_{\text{conf}} \\
        c p_i ,& \text{otherwise},
    \end{cases}
\end{equation}
where $c$ is the objectness confidence predicted by the model and $p_i$ is the probability of the $i$-th class. \Cref{eq:conf} maximizes the probability of each class by increasing the objectness confidence while the probabilities for less possible classes are raising as well to accelerate search speed since multiple labels for an object are allowed in an object detection task. As a result, the total objects fed into NMS can be increased.

\begin{figure}[t]
\centering
    \includegraphics[trim={4.8cm 1.0cm 5.1cm 1.3cm},clip,width=0.8\linewidth]{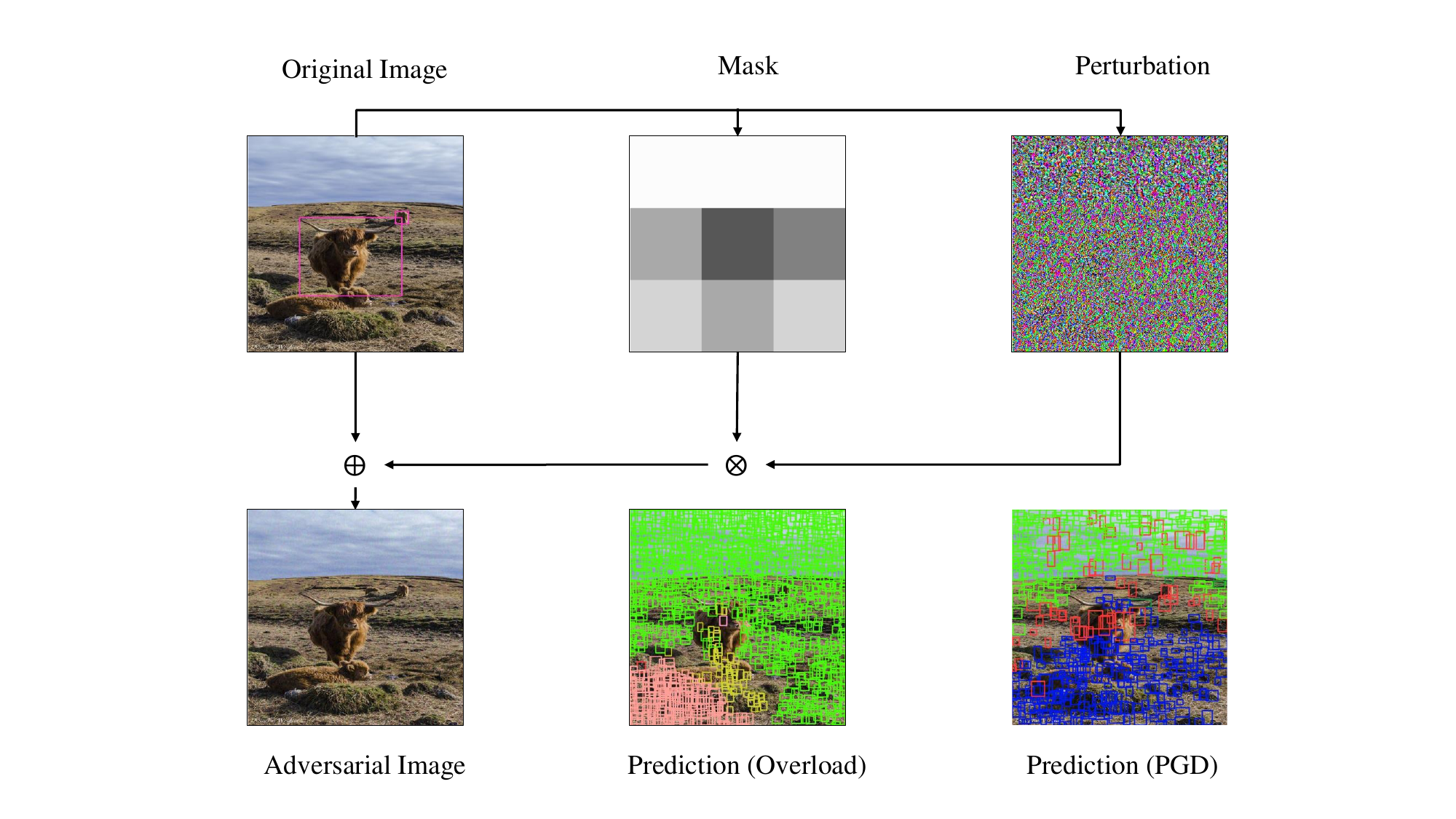}
    \caption{The execution flow of spatial attention.}
    \label{fig:box_algo}
\end{figure}

We introduce a novel approach, called spatial attention, to enhance the efficacy of latency attacks. The purpose of spatial attention is to force the attacking processing to pay more attention to the low-density regions rather than the regions that already have objects. The flow of spatial attention is visualized in \Cref{fig:box_algo}. Throughout each iteration for crafting adversarial examples, the weight assigned to the central area is minimized as it contains two objects. Conversely, boundaries, which are either empty or overlaid with objects, have their corresponding weights adjusted accordingly.

We summarize the steps of our method below. First, the image is divided into an $m\times m$ grid. Second, the weight of each grid cell, denoted as $W_{i,j}$, is adjusted based on the number of objects obtained from $M(x)$. The weight of the particular grid cell is decreased if adversarial attacks cannot produce more objects in the area or if it is dense enough. Finally, the adversarial image is updated using the following equation:
\begin{equation}
\label{eq:pgd_sa}
    x_{i,j}^{\text{adv}} = x_{i,j}^{\text{org}} + \eta W_{i,j} \Delta x_{i, j} 
\end{equation}
where $i$ and $j$ are grid indices; $x^{\text{adv}}$ is the adversarial image; $x^{\text{org}}$ is the original image; $\eta$ is step size, and $\Delta x$ is the modification of the image. Here $\Delta x = \nabla_x L$, the gradient for the loss function specified in \Cref{eq:loss_fun}, where
$$L=\sum^{n}_{i=0} F_l(F_{\text{conf}}(M(x)_i)).$$

\section{Experiments}
\label{sec:exp}

\subsection{Setup}
To demonstrate the potential impacts of latency attacks on systems with limited computing resources, we conducted experiments on Nvidia Jetson NX, a popular edge device for real-time object detection applications \cite{ravi2022real, liang2022edge}, and the used models are YOLOv5s and YOLOv5n due to their low computational requirements. To further accelerate the running speed, the inference models are compiled into the TensorRT format \cite{jeong2022tensorrt}. We randomly selected 1,000 images from the validation set of MS COCO 2017 dataset \cite{lin2014microsoft}. Meanwhile, the batch size is 1 and the images' dimension is $(640, 640)$. The results are sorted by the elapsed time and presented in the selected percentile. Additionally, we have included ablation studies in the Appendix, which covers various aspects such as the selection of the loss function, evaluation of spatial attention, transfer attack analysis, and the performance impact on different models. The results obtained indicate that the total number of objects generated by our proposed attack closely aligns with the theoretical maximum that victim models can output, demonstrating the effectiveness of our approach. Furthermore, our method allows the encoding of information from multiple models within a single image, showcasing the feasibility of an ensemble attack to deceive various object detection models.

\subsection{Latency Evaluation}

\begin{table*}[t]
    \centering
    \begin{tabular}{ccccccc|cccccc}
        \toprule
        \multirow{3}{*}{Percentile} & \multicolumn{6}{c|}{YOLOv5s} & \multicolumn{6}{|c}{YOLOv5n} \\ 
        \cline{2-13}
        & \multicolumn{3}{c}{Adversarial Examples}  & \multicolumn{3}{c|}{Original Examples}
        & \multicolumn{3}{|c}{Adversarial Examples} & \multicolumn{3}{c}{Original Examples} \\
        & objects & boxes & time & objects & boxes & time & objects & boxes & time & objects & boxes & time \\
        \midrule
        min    &  4098 & 1587 &  18.7 &    0 &  0 & 14.1   &     5893 & 1131 &  14.7 &  0 & 0 &  9.3 \\
        0.10   & 23112 & 2034 & 122.2 &   40 &  3 & 16.4   &    11925 & 1421 &  22.6 &  0 & 0 & 10.7 \\
        0.25   & 24248 & 4653 & 179.6 &   47 &  4 & 16.4   &    16405 & 2303 &  39.8 &  6 & 1 & 11.5 \\
        0.50   & 24000 & 2250 & 217.8 &   28 &  3 & 16.4   &    22363 & 2315 & 128.9 & 22 & 4 & 11.5 \\
        0.75   & 24335 & 4514 & 264.2 &    3 &  1 & 16.6   &    20923 & 2372 & 208.3 & 97 & 8 & 11.5 \\
        0.90   & 24778 & 2340 & 278.4 &  234 & 21 & 16.7   &    21947 & 2659 & 226.7 &  6 & 2 & 11.6 \\
        max    & 24859 & 2363 & 293.8 &  212 & 21 & 16.9   &    23396 & 2502 & 258.7 & 25 & 2 & 12.0 \\
        \bottomrule
    \end{tabular}
    \caption{Results of latency evaluation on NVIDIA Jetson NX}
    \label{table:nms_nx}
\end{table*}

\begin{table}[t]
    \centering
    \begin{tabular}{ccc|cc}
        \toprule
        \multirow{2}{*}{Ratio} & \multicolumn{2}{c|}{YOLOv5s} & \multicolumn{2}{|c}{YOLOv5n}  \\ 
        \cline{2-5}
        & time [ms] & FPS & time [ms] & FPS\\
        \midrule
        0.00   &   23 & 43.5 &   20 & 50.0 \\
        0.25   &   81 & 12.3 &   56 & 17.8 \\
        0.50   &  136 &  7.3 &   92 & 10.8 \\
        0.75   &  180 &  5.5 &  114 &  8.8 \\
        1.00   &  230 &  4.3 &  135 &  7.4 \\
        \bottomrule
    \end{tabular}
    \caption{Pipeline simulator on NVIDIA Jetson NX}
    \label{table:pipeline_nx}
\end{table}

\begin{table}[t]
    \centering
    \begin{tabular}{c|cccc}
     GPU model     & $T_{org}$  [ms] &  $T_{atk}$ [ms] \\
    \midrule
    Nvidia A100    &  16  &  54 \\
    Nvidia A10     &  14  &  52 \\
    Nvidia 1080 Ti &  11  &  62 \\
    Nvidia Jetson  &  23  & 230 \\
    \end{tabular}
    \caption{Latency evaluation across various GPUs}
    \label{table:latency_vs_gpus}
\end{table}

\Cref{table:nms_nx} presents the experimental results of YOLOv5s and YOLOv5n on Nvidia Jetson NX, where \textit{objects} mean the total number of objects fed into NMS; \textit{boxes} mean the total number of bounding boxes output by NMS, and \textit{time} is the elapsed time in millisecond, including model inference and NMS.

The elapsed times of the original examples with YOLOv5s and YOLOv5n are about 16.4 ms and 11.5 ms respectively for 50\% percentile. Conversely, the average elapsed times of adversarial examples are about 13.0$\times$ and 11.0$\times$ longer than that of the original examples with YOLOv5s and YOLOv5n respectively. Under our experimental configuration, the maximum number of objects predicted by the YOLO model is 25,200. Results also demonstrate that our attacks can generate over 20,000 ghost objects for most images: over 90\% of images for YOLOv5s and over 60\% of images for YOLOv5n.

To evaluate the impact of the latency attack on real-world applications, we measured the elapsed time of 1,000 images processed by a pipeline. \Cref{table:pipeline_nx} shows the average execution time of each image in milliseconds, where the Ratio column represents the proportion of adversarial images in the test dataset, and FPS is the abbreviation for "Frame Per Second". As the proportion of adversarial images grows, the elapsed time rises rapidly and the inference time increases tenfold in the worst case. The results suggest that our proposed latency attack poses a potential warning in real-world applications.

We conducted the same pipeline experiment with YOLOv5s across various GPUs. \Cref{table:latency_vs_gpus} presents the experimental results, where $T_{org}$ is measured without the involvement of adversarial examples; $T_{atk}$ refers to the scenario where all inputs are adversarial examples. As can be seen, the ratio of the elapsed time of $T_{atk}$ to $T_{org}$ is about 3-6 (excluding Nvidia Jetson). These findings suggest that the attack can be generalized across different GPUs, but the strength of the proposed attack depends on hardware configurations.

\subsection{IoU-invariant Verification}
\begin{table*}[t]
    \centering
    \begin{tabular}{cccc|ccc|ccc|ccc}
        \toprule
        \multirow{3}{*}{Percentile} & \multicolumn{12}{c}{YOLOv5s} \\ 
        \cline{2-13}
        & \multicolumn{3}{c}{$T_{\text{IoU}}$=0.00} & \multicolumn{3}{c}{$T_{\text{IoU}}$=0.45}
        & \multicolumn{3}{c}{$T_{\text{IoU}}$=0.80} & \multicolumn{3}{c}{$T_{\text{IoU}}$=1.00} \\
        & objects & boxes & time & objects & boxes & time & objects & boxes & time & objects & boxes & time \\
        \midrule
        min    &  4098 & 403 &  17.3   &   4098 & 1587 &  18.7   &    4098 &  2198 &  17.9 &    4098 &  4098 &  19.3 \\
        0.10   & 23635 & 623 &  64.8   &  23112 & 2034 & 122.2   &   20157 &  6352 & 156.9 &   20328 & 20328 & 125.7 \\
        0.25   & 24027 & 745 & 120.2   &  24248 & 4653 & 179.6   &   23345 &  7756 & 211.5 &   23476 & 23476 & 215.8 \\
        0.50   & 20579 & 550 & 200.0   &  24000 & 2250 & 217.8   &   24345 & 12504 & 234.7 &   24381 & 24381 & 245.4 \\
        0.75   & 24264 & 692 & 265.6   &  24335 & 4514 & 264.2   &   24667 &  7544 & 248.0 &   24138 & 24138 & 275.5 \\
        0.90   & 24268 & 677 & 274.5   &  24778 & 2340 & 278.4   &   24338 & 11593 & 260.4 &   23356 & 23356 & 280.2 \\
        max    & 24810 & 476 & 289.2   &  24859 & 2363 & 293.8   &   24859 &  8039 & 269.7 &   24905 & 24905 & 285.5 \\
        \bottomrule
    \end{tabular}
    \begin{tabular}{cccc|ccc|ccc|ccc}
        \toprule
        \multirow{3}{*}{Percentile} & \multicolumn{12}{c}{YOLOv5n} \\ 
        \cline{2-13}
        & \multicolumn{3}{c}{$T_{\text{IoU}}$=0.00} & \multicolumn{3}{c}{$T_{\text{IoU}}$=0.45}
        & \multicolumn{3}{c}{$T_{\text{IoU}}$=0.80} & \multicolumn{3}{c}{$T_{\text{IoU}}$=1.00} \\
        & objects & boxes & time & objects & boxes & time & objects & boxes & time & objects & boxes & time \\
        \midrule
        min    &  5893 & 393 &  14.3   &   5893 & 1131 &  14.7   &     5893 &  2429 &  15.2 &    5893 &  5893 &  15.3 \\
        0.10   & 11660 & 533 &  21.9   &  11925 & 1421 &  22.6   &    11904 &  4580 &  24.1 &   11961 & 11961 &  25.4 \\
        0.25   & 16858 & 529 &  31.3   &  16405 & 2303 &  39.8   &    16680 &  5008 &  44.4 &   14789 & 14789 &  48.3 \\
        0.50   & 21423 & 660 &  52.1   &  22363 & 2315 & 128.9   &    21440 & 10020 &  60.5 &   19231 & 19231 & 139.2 \\
        0.75   & 22905 & 540 &  62.6   &  20923 & 2372 & 208.3   &    21876 & 11360 & 152.4 &   21748 & 21748 & 210.8 \\
        0.90   & 21536 & 562 & 163.2   &  21947 & 2659 & 226.7   &    22279 & 10305 & 223.7 &   21796 & 21796 & 225.9 \\
        max    & 22897 & 538 & 234.7   &  23396 & 2502 & 258.7   &    22959 &  9505 & 241.1 &   23396 & 23396 & 256.2 \\
        \bottomrule
    \end{tabular}
    \caption{The elapsed time with different IoU threshold on NVIDIA Jetson NX}
    \label{table:nms_iou_nx_5}
\end{table*}

This experiment investigates how the number of surviving objects influences execution time. The results are presented in \Cref{table:nms_iou_nx_5}, where $T_{\text{IoU}}$ represents the IoU threshold. $T_{\text{IoU}}=1.0$ and $T_{\text{IoU}}=0.0$ simulate the worst and best cases, respectively. $T_{\text{IoU}}=0.45$ represents the default configuration, and $T_{\text{IoU}}=0.80$ is an intermediate case.

As indicated in \Cref{table:nms_iou_nx_5}, our attack produces approximately 550 and 600 non-overlapping boxes for YOLOv5s and YOLOv5n, respectively, at the 50\% percentile. Although the best cases reduce over 95\% of objects, the corresponding elapsed times are quite similar to those of the other cases. When compared to the results obtained from synthetic data, as illustrated in \Cref{fig:obj_nms_nx,} the latency increments for real-world data are marginal, owing to the variability in the strength of adversarial examples. Producing more survival boxes to increase execution time is only recommended for black-box scenarios or situations where the number of objects is saturated.
\subsection{Comparison to Known Attacks}
\begin{table}[t]
    \centering
    \begin{tabular}{c|cccc}
     Paper   & $\epsilon$ &  Objects  & Recall  & mAP@50\\
    \midrule
    Phantom Sponge  & 70  & 19000 & 18  & - \\
    Phantom Sponge  & 30  &  9000 & 77  & - \\
    Overload        & 15  & 20000 & 44  & $<$0.001 \\
    Daedalus        & -   & ~1500 &  -  & $<$0.001 \\
    \end{tabular}
    \caption{Performance comparison with existing works}
    \label{table:comp}
\end{table}

\begin{table}[t]
    \centering
    \begin{tabular}{c|ccc}
     Method   & BatchSize  &  Memory (GB) & Time (s) \\
    \midrule
    Overload         & 16  & 7.5 & 180 \\
    Overload + $L_{IoU}$   & 16  &  -  & - \\
    Overload + $L_{area}$  & 16  & 7.9 & 240 \\
    \end{tabular}
    \caption{Resource Consumption Comparison for Overload with Different Losses}
    \label{table:comp_with_different_loss}
\end{table}
Phantom Sponge \cite{shapira2023phantom} and Daedalus \cite{wang2021daedalus} generate numerous objects by reducing the bounding box sizes, implicitly minimizing pairwise IoU scores. This operation is equivalent to maximizing the term $|\mathcal{B}|$ in \Cref{eq:T_cost}. However, they necessitate tracking the gradient of IoU computations, leading to high memory usage and longer processing times without inducing performance gains. In contrast, the proposed method simplifies the objective, conserving computational time.

\Cref{table:comp} compares Overload to existing works, where $\epsilon$ is the maximum size of the perturbation; Objects represent the total number of objects; Recall measures the similarity of predictions between clean and adversarial images; mAP@50 is the mean average precision calculated at IoU threshold 0.5; and "-" refers to metrics that are not applicable. For the L2-norm attack, Daedalus produces fewer objects than other competitors. Meanwhile, Phantom Sponge generates about 19,000 objects within a perturbation size $\epsilon=70.0$, but our attack produces more objects and higher recall within $\epsilon=15.0$. These comparison results should not be over-interpreted, as the goals and constraints on perturbations for Phantom Sponge and Daedalus differ from ours.

We conducted experiments on Nvidia V100 (32GB) to quantify memory usage and the total time required to craft adversarial examples. The gradients of tracking IoU calculation for all objects require more memory than is available in Nvidia V100. Consequently, Daedalus tracks partial objects to mitigate memory usage. Crafting 10 adversarial examples sequentially with Daedalus takes approximately 145 minutes, with a peak memory usage of around 8.5GB. In contrast, our method can process 16 images in a batch in about 3 minutes, with a peak memory usage of 7.5GB. Phantom Sponge limits the number of objects used to compute the IoU score to less than 100, resulting in slightly higher resource usage than our method under the same conditions. However, Phantom Sponge needs to execute NMS several times in the attacking processing, making a longer execution time when the total objects grow.

\diff{
\Cref{table:comp_with_different_loss} represents the evaluation of Overload integrated with an additional loss, where "Overload + $L_{IoU}$" refers to the objective involving minimizing pairwise IoU scores, and "Overload + $L_{area}$" refers to the objective involving minimizing the areas of objects. "Overload + $L_{IoU}$" failed to execute due to out-of-memory issues. The resource requirements for "Overload + $L_{area}$" are higher than native Overload. We attempted various function formulations for minimizing areas and coefficients to amplify their influence. However, we obtained the same performance as those presented in IoU-invariant verification. Therefore, we believe that Overload outperforms these works in terms of the number of objects without needing a hyperparameter for confidence-IoU balance, while using fewer computing resources and simplifying the objective design.
}

\subsection{Discussion}
In spirit, our proposed latency attack can be viewed as a variant of the appearing attack, but these two attacks have fundamentally different purposes. The original appearing attack \cite{yin2022adc} creates a specific type of object such that the detector makes an unexpected decision. On the contrary, the latency attack produces numerous objects and the detector cannot process those objects within the strict time restriction. Nevertheless, it is important to acknowledge certain limitations and potential impacts associated with our work, which are addressed below.

\textbf{NMS-agnosticity} We argue that our attack is a generic attack against NMS. The objective of the latency attack is to maximize the confidence of individual object without any information about locations. DIoU \cite{zheng2020distance}, CIoU \cite{zheng2020distance}, and others implementations refine the pruning criterion or scoring function. Those modifications do not affect the time complexity. Hence, any object detection task that requires NMS as post-processing is a potential victim of our latency attack.

\textbf{Attack scenarios} This kind of adversarial image requires more computing resources than normal images during the inference time. This implies it can be leveraged as a Denial-of-Service (DoS) attack and resources may be exhausted in shared-resource environments. Furthermore, devices with limited computing resources are potential attack candidates. Although some models implement a timeout mechanism, once the execution time is exhausted after processing any example in the batch, the rest of the examples in the batch are skipped. Additionally, issuing the next batch is delayed until the previous processing is complete. This property could be  exploited maliciously.

The ghost objects generated by our latency attack could increase processing time or have unforeseen consequences for downstream tasks. For instance, the collision avoidance system identifies risky objects and predicts the trajectories of those objects \cite{li2021driver}. The execution time worsens if the predictions of an adversarial image are fed into the collision avoidance system.

\textbf{Limitation} Our evaluation of this attack is under white-box settings. We acknowledge that conducting adversarial attacks in this setting may seem impractical. However, they can serve as motivation to refine more robust model designs. Generating a physically universal patch that can increase inference time under black-box settings represents a critical area of investigation. Nonetheless, evaluating practicality falls beyond the scope of this work.

While our attack cannot be directly applied to NMS-free models like DETR \cite{carion2020end} and CenterNet \cite{zhou2019objects}, it's vital not to disregard the potential impact on these models. However, a recent benchmark \cite{liu2021dataset} evaluated popular underwater object detection models on edge devices, revealing that 11 out of 12 models still rely on NMS. Therefore, our work is important in drawing attention to the potential risks in object detection and inspiring researchers to explore mitigation techniques applicable across various applications.

\textbf{Defenses} There are three probable defenses. The first one limits the maximum execution time. However, the predefined threshold should be reconfigured when the model is deployed on different hardware. Besides, the previous discussion has shown the risks of timeouts. The second one limits the total number of objects fed into NMS. The choice of the appropriate value depends on the used model and the specific dataset. A strict limitation can mitigate the latency issue but it may cause some objects to be ignored by the model. The last one directly strengthens the reliability of models with adversarial training and other defensive strategies.

\section{Conclusion}
In this paper, we proposed a novel latency attack, called Overload, by targeting the NMS module in object detection tasks. To fully exploit the vulnerabilities, we have analyzed the time complexity of NMS on GPUs, and observed that the execution time is dominated by the number of objects fed into NMS. Based on the analysis, we proposed a new formulation to craft adversarial images for latency attacks, and a new technique, called spatial attention, to activate more objects in particular areas to improve attack efficiency. Compared with existing works, Overload achieves superior attack performance but with smaller computational costs and memory usage. With the attack, the execution time of a single adversarial image is about ten times longer than that of a normal image on Nvidia Jetson NX. It poses a significant threat to systems with limited computing resources. Furthermore, we establish the NMS-agnostic nature of our proposed attack, highlighting its potential to become a common vulnerability to all object detection systems that rely on NMS.

There are many future directions to explore. First, the black-box latency attacks have not been studied yet. Spatial attention is a gradient-free procedure, so it may be a useful tool for black-box attacks. Second, the performance impact of the latency attack for downstream tasks needs more investigation. Lastly, developing defense strategies against latency attacks is an important topic but beyond the scope of this paper. We acknowledge that our findings might be used as exploits. However we believe our disclosure of vulnerability analysis is necessary to accelerate the design of more resilient and safe edge devices against latency attacks.

\subsection*{Acknowledgments.}
This research is partially supported by NTSC, Taiwan, R.O.C. under Grant no. 112-2634-F-007-002. We would also like to thank to National Center for High-performance Computing (NCHC) for providing computational and storage resources.

{
    \small
    \bibliographystyle{ieeenat_fullname}
    \bibliography{bib_refs}
}

\appendix
\section{Comprehensive Analysis of NMS Algorithm}
\label{sec:nms_algo}
As described in earlier section, the steps of NMS can be divided into four major tasks:
\begin{enumerate}
    \item Filtering low-confidence objects;
    \item Sorting candidates by probabilities;
    \item Calculating pairwise IoU scores;
    \item Pruning inactive objects.
\end{enumerate}

Let $\mathcal{C}$ be the set of objects fed into NMS. In this first step, the filtering step scans all objects in the set and deletes low-confidence objects, resulting in linear time complexity $O(|\mathcal{C}|$. In the second step, the time complexity of sorting is well known as $O(|\mathcal{C}|\log(|\mathcal{C}|))$. In the third step, NMS constructs a $|\mathcal{C}| \times |\mathcal{C}|$ matrix which stores the pairwise comparison results. IoU score computes the area of overlap between a pair of selected objects requiring a constant number of float operations. Therefore, the time complexity is $O(|\mathcal{C}| \times |\mathcal{C}|)$. In the last step, the major goal is to prune unqualified objects based on the IoU scores. The details are listed in Algorithm \ref{algo:nms_pruning}, where $\mathbf{S}_{|\mathcal{C}| \times |\mathcal{C}|}$ stores IoU scores. An object is marked duplicated if the IoU score $\mathbf{S}[i][j]$ is greater than the NMS threshold $N_t$. $\mathbf{r}$ is a utility array tacking whether the objects are marked. To obtain the worst case, we assume that no objects are duplicated, making the condition in line \ref{algo:nms_pruning:cond} always satisfied. As a result, the algorithm can be simplified to a two-layer loop that traverses half elements in the matrix $\mathbf{S}$. Therefore, the overall time complexity is independent of the rate of survival objects or the properties of boxes although some elements are marked removable during the pruning procedure.

\begin{algorithm}[t]
\caption{NMS Pruning}
\label{algo:nms_pruning}
\begin{algorithmic}[1]
\STATE {$n \leftarrow |\mathcal{C}|$}
\STATE {$\mathbf{r}=\mathbf{0}$}
\STATE {$\mathcal{D} \leftarrow \{\}$}
\FOR{$i = 1 \text{ to } n$}
    \IF{$\mathbf{r}[i] \neq \text{True}$}
    \label{algo:nms_pruning:cond}
        \STATE {$\mathcal{D} \leftarrow \mathcal{D} \cup \{i\}$}
        \FOR{$j = i + 1 \text{ to } n$}
            \STATE {$\mathbf{r}[j] = \text{bool}(\mathbf{r}[j] + \mathbf{S}[i][j] > N_t)$}
        \ENDFOR
    \ENDIF
\ENDFOR
\end{algorithmic}
\end{algorithm}

\section{Ablation Study}

\subsection{Impact of Different Objective Functions}
\begin{table*}[ht]
    \centering
    \begin{tabular}{cccc|ccc|ccc|ccc}
        \toprule
        \multirow{3}{*}{Percentile} & \multicolumn{12}{c}{YOLOv5s}\\ 
        \cline{2-13}
        & \multicolumn{3}{c}{$F=\log(x)$}  & \multicolumn{3}{c|}{$F=\text{tanh}(x)$}
        & \multicolumn{3}{|c}{$F=x^2/2$} & \multicolumn{3}{c}{$F=-\log(1-x)$} \\
        & objects & boxes & time & objects & boxes & time & objects & boxes & time & objects & boxes & time\\
        \midrule
        min    &  3228  &  392 &  52.8 &  9269 & 1081 &  25.0 &  3035 & 228 & 17.8 &  4556 & 511 & 19.3 \\
        0.10   & 18900  & 1570 &  52.8 & 19546 & 1829 &  45.9 &  5708 & 357 & 19.6 &  9553 & 555 & 23.9 \\
        0.25   & 23494  & 2817 &  91.8 & 17140 & 1471 & 148.3 &  6644 & 427 & 20.4 & 10079 & 647 & 25.0 \\
        0.50   & 22273  & 2617 & 167.4 & 19677 & 1961 & 192.1 &  6934 & 413 & 21.4 & 10955 & 579 & 26.6 \\
        0.75   & 22660  & 1630 & 208.5 & 21559 & 1448 & 227.1 &  8681 & 502 & 23.0 & 11529 & 644 & 27.9 \\
        0.90   & 22951  & 3162 & 241.2 & 22231 & 2585 & 241.6 & 10617 & 548 & 26.0 & 12422 & 790 & 29.1 \\
        max    & 23922  & 1386 & 241.2 & 22944 & 1382 & 253.0 & 13104 & 824 & 30.2 & 12632 & 813 & 39.7 \\
        \bottomrule
    \end{tabular}

    \begin{tabular}{cccc|ccc|ccc|ccc}
        \toprule
        \multirow{3}{*}{Percentile} & \multicolumn{12}{c}{YOLOv5n}\\ 
        \cline{2-13}
        & \multicolumn{3}{c}{$F=\log(x)$}  & \multicolumn{3}{c|}{$F=\text{tanh}(x)$}
        & \multicolumn{3}{|c}{$F=x^2/2$} & \multicolumn{3}{c}{$F=-\log(1-x)$} \\
        & objects & boxes & time & objects & boxes & time & objects & boxes & time & objects & boxes & time\\
        \midrule
        min    &  2890  &  763 &  13.3 &  4134 &  579 &  14.3 & 3435 & 210 & 13.0 &  3060 & 294 & 12.9 \\
        0.10   &  9012  & 1145 &  19.5 & 11950 & 1103 &  23.9 & 5295 & 339 & 14.3 &  8009 & 573 & 17.2 \\
        0.25   & 12985  & 1474 &  26.0 & 14064 & 1194 &  27.6 & 6074 & 355 & 15.0 &  8751 & 600 & 18.2 \\
        0.50   & 16755  & 1737 &  34.3 & 16128 & 1319 &  32.0 & 6920 & 444 & 15.8 &  9631 & 697 & 19.5 \\
        0.75   & 18910  & 1766 &  69.7 & 18276 & 1206 &  37.2 & 8079 & 448 & 17.1 & 10909 & 656 & 21.2 \\
        0.90   & 20901  & 1649 & 135.5 & 17507 & 1460 & 117.9 & 9361 & 612 & 18.7 & 11746 & 684 & 22.6 \\
        max    & 22303  & 1472 & 231.6 & 20902 & 1238 & 206.7 & 9946 & 585 & 25.3 & 15479 & 983 & 30.4 \\
        \bottomrule
    \end{tabular}

    \caption{The total elapsed time using different objective function on NVIDIA Jetson NX}
    \label{table:nms_loss_nx_5}
    
\end{table*}

In Section \ref{sec:algo}, we mentioned that any monotonic increasing function can be used as the loss function. In this experiment, we evaluated the performance of four qualified functions: $\log(x)$, $\tanh(x)$, $x^2/2$, and $-\log(1-x)$. $\log(x)$. The derivative of $\tanh(x)$ is $1$ at $x=0$ and smoothly decreases as $x$ increases. $x^2/2$ is a convex function with its minimum point at $x=0$, and its derivative gradually increases. The limit of $-\log(1-x)$ as $x$ approaches $1$ from the left is positive infinity.

Table \ref{table:nms_loss_nx_5} shows that $x^2/2$ and $-\log(1-x)$ result in significantly fewer total objects compared to the other functions. To explain this phenomenon, we divide the objects predicted by the model into two sets: $\mathcal{S}^+$ and $\mathcal{S}^-$, where $\mathcal{S}^+ = \{x | \text{conf}(M(x)i) > T{\text{IoU}}\}$ and $\mathcal{S}^- = \{x | \text{conf}(M(x)i) \leq T{\text{IoU}}\}$. The original objective of the loss function is to maximize the confidence of individual boxes in set $\mathcal{S}^-$, increasing the total number of objects fed into NMS. However, $-\log(1-x)$ tends to increase the confidence of boxes in set $\mathcal{S}^-$ due to the rapid growth of its derivative when $x$ is close to $1.0$. As a result, although PGD maximizes the objective function defined in (\ref{eq:loss_fun}), $-\log(1-x)$ yields the worst performance. The behavior of $x^2/2$ is similar, as its derivative increases gradually. Therefore, an effective objective function should not only be a monotonic increasing function but also have a monotonically decreasing derivative.

\subsection{Spatial Attention Evaluation}
This experiment evaluates the influence of spatial attention. Table \ref{table:sa} shows the experimental results on Nvidia Jetson NX, where \textit{SA} means the adversarial attack with the spatial attention and \textit{PGD} means the native implementation of PGD. As can be seen, one can find that the spatial attention method can generate approximately 2,000 or more objects for the YOLOv5s and YOLOv5n models. This increase in object count is due to the iterative generation of objects from regions with fewer objects, facilitated by the proposed spatial attention technique.

Table \ref{table:sa_grid} offers a performance comparison with different grid sizes, where \textit{Grid} signifies that the image is tiled into $k\times k$ grids. In the case of $1\times1$ tiling, the configuration reverts to the original PGD attack, where all pixels are part of the same grid, sharing identical weights. Upon introducing spatial attention, specific areas can be highlighted, resulting in a performance gain. However, when the image is divided into a $20\times20$ grid, some tiles remain unhighlighted, hindering effective attacks on those tiles. The experimental findings suggest that an optimal configuration for spatial attention is around $5\times5$.

These results demonstrate the effectiveness of the spatial attention technique in generating a larger number of objects. The comparison between SA and PGD sheds light on the potential vulnerabilities and limitations of the YOLOv5 models when exposed to adversarial attacks with spatial attention.

\begin{table*}[tb]
    \centering
    \begin{tabular}{ccccccc|cccccc}
        \toprule
        \multirow{3}{*}{Percentile} & \multicolumn{6}{c|}{YOLOv5s} & \multicolumn{6}{|c}{YOLOv5n} \\ 
        \cline{2-13}
        & \multicolumn{3}{c}{SA}  & \multicolumn{3}{c|}{PGD}
        & \multicolumn{3}{|c}{SA} & \multicolumn{3}{c}{PGD} \\
        & objects & boxes & time & objects & boxes & time & objects & boxes & time & objects & boxes & time \\
        \midrule
        min    &  4098 & 1587 &  18.7 &   3228 &  392 &  18.3 &     5893 & 1131 &  14.7 &  2890 &  763 &  14.8 \\
        0.10   & 23112 & 2034 & 122.2 &  18900 & 1570 & 117.8 &    11925 & 1421 &  22.6 &  9012 & 1145 &  27.1 \\
        0.25   & 24248 & 4653 & 179.6 &  23494 & 2817 & 161.9 &    16405 & 2303 &  39.8 & 12985 & 1474 &  35.9 \\
        0.50   & 24000 & 2250 & 217.8 &  22273 & 2617 & 209.6 &    22363 & 2315 & 128.9 & 16755 & 1737 &  59.3 \\
        0.75   & 24335 & 4514 & 264.2 &  22660 & 1630 & 232.2 &    20923 & 2372 & 208.3 & 18910 & 1766 & 175.2 \\
        0.90   & 24778 & 2340 & 278.4 &  22951 & 3162 & 246.2 &    21947 & 2659 & 226.7 & 20901 & 1649 & 210.8 \\
        max    & 24859 & 2363 & 293.8 &  23922 & 1386 & 270.5 &    23396 & 2502 & 258.7 & 22303 & 1472 & 244.6 \\
        \bottomrule
    \end{tabular}
    \caption{Results of spatial attention evaluation}
    \label{table:sa}
\end{table*}

\begin{table*}[tb]
    \centering
    \begin{tabular}{ccccccc|cccccc}
        \toprule
        \multirow{3}{*}{Percentile} & \multicolumn{6}{c|}{YOLOv5s} & \multicolumn{6}{|c}{YOLOv5n} \\ 
        \cline{2-13}
        & \multicolumn{2}{c}{Grid=$1\times1$}  & \multicolumn{2}{c}{Grid=$5\times5$} & \multicolumn{2}{c|}{Grid=$20\times20$}
        & \multicolumn{2}{|c}{Grid=$1\times1$} & \multicolumn{2}{c}{Grid=$5\times5$} & \multicolumn{2}{c}{Grid=$20\times20$} \\
        & objects & boxes & objects & boxes & objects & boxes & objects & boxes & objects & boxes & objects & boxes \\
        \midrule
        min    &  3421 &  507 &  7465 &  1235 &  6220 & 1062 &     3421 &  507 &  7465 & 1235 &  6220 & 1062 \\
        0.10   &  6359 &  983 & 10739 &  1405 &  8193 & 1412 &     6359 &  983 & 10739 & 1405 &  8193 & 1412 \\
        0.25   & 10696 & 1203 & 11197 &  1535 &  9706 & 1595 &    10596 & 1203 & 11197 & 1535 &  9706 & 1595 \\
        0.50   & 12557 & 1485 & 16769 &  1770 & 21520 & 1827 &    12557 & 1485 & 16768 & 1770 & 21520 & 1827 \\
        0.75   & 17671 & 1848 & 17461 &  2110 & 19781 & 2117 &    17671 & 1848 & 17461 & 2110 & 19781 & 2117 \\
        0.90   & 20647 & 2155 & 17551 &  2305 & 18491 & 2315 &    20647 & 2155 & 17551 & 2305 & 18491 & 2315 \\
        max    & 21515 & 2723 & 20202 &  2688 & 19949 & 2638 &    21514 & 2723 & 20202 & 2688 & 19949 & 2638 \\
        \bottomrule
    \end{tabular}
    \caption{Performance comparison with different grid size.}
    \label{table:sa_grid}
\end{table*}

\section{Latency Attacks on Various Models}
This experiment aims to comprehensively evaluate the performance of the proposed latency attack on various models. To extend our analysis, we conducted additional experiments on YOLOv3 model and two latest YOLO models, namely YOLOv7 and YOLOv8. Tables \ref{table:nms_nx_yolo} presents the results obtained from these models, where the elapsed time was not measured as these models are not fully optimized for edge devices.

The obtained results reveal the effectiveness of our attack. At the 50th percentile, our attack generates 20,578 objects for YOLOv7 and 23,163 objects for YOLOv7-tiny. Similarly, for YOLOv8, our attack generates 8,313 objects at the 50th percentile. These numbers are in comparison to the maximum number of objects predicted by YOLOv7 (25,200 objects) and YOLOv8 (8,400 objects). The significant number of objects generated by our attack demonstrates its potency. Consequently, our findings indicate that both YOLOv7 and YOLOv8 models are susceptible to latency attacks. It is worth mentioning that as the total number of objects increases, the elapsed time during the attack is expected to rise.

We argue that object detection models with different architectures are also vulnerable to latency attacks.  However, it is important to note that the proposed objective in (\ref{eq:loss_fun}) is specifically designed for YOLO series and may not be the optimal objective for other architectures. To investigate this further, we conducted a small experiment using the SSD model \cite{liu2016ssd}. Our observations revealed that the number of objects generated using our attack ranged from approximately 300 to 1,000. Since SSD normalizes the probabilities using the softmax function, ensuring that the sum of probabilities for all classes is 1.0, the attack's performance is highly influenced by the image's characteristics and the target class. These findings suggest that SSD models are indeed vulnerable to latency attacks, but further improvements can be made.

\begin{figure}[t]
    \centering
    \begin{subfigure}{.45\linewidth}
        \centering
        \includegraphics[width=1.0\linewidth]{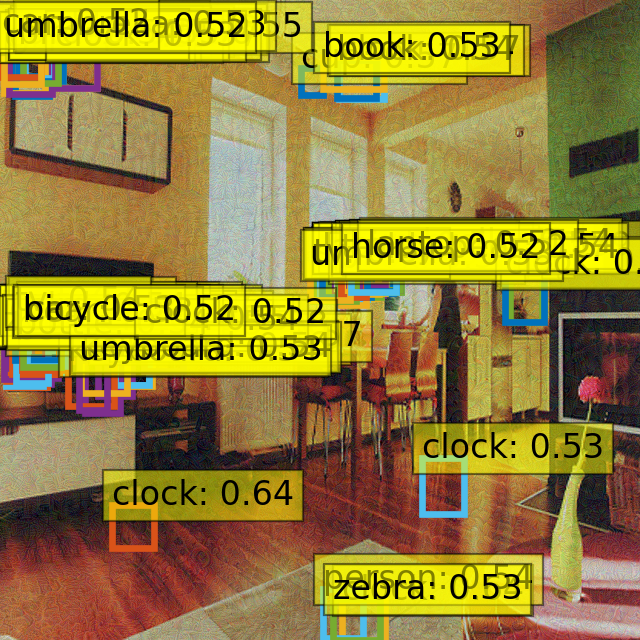}
        \caption{}
        \label{fig:example_retinanet_raw}
    \end{subfigure}
    \begin{subfigure}{0.45\linewidth}
        \centering
        \includegraphics[width=1.0\linewidth]{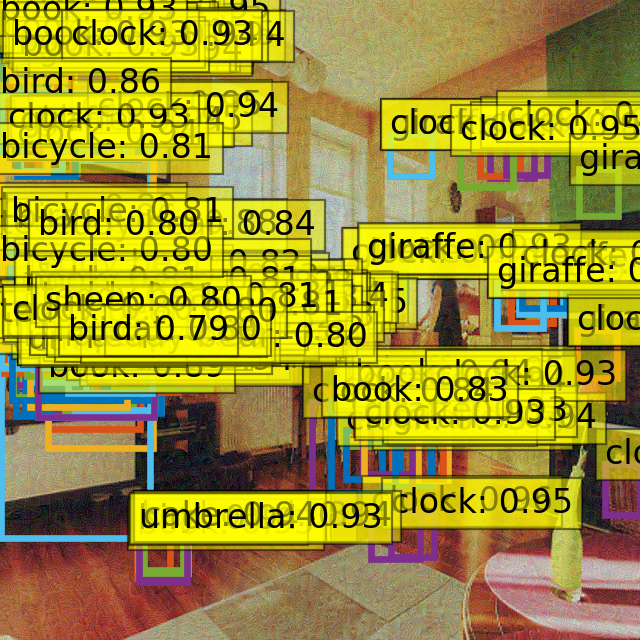}
        \caption{}
        \label{fig:example_retinanet_overload}
    \end{subfigure}
    \caption{The outputs of the adversarial examples by Retinanet. \ref{fig:example_retinanet_raw} and \ref{fig:example_retinanet_overload} are generated by the normal PGD attack and Overload attack, respectively. }
    \label{fig:example_retinanet}
\end{figure}

\begin{figure}[t]
    \centering
    \begin{subfigure}{.45\linewidth}
        \centering
        \includegraphics[width=1.0\linewidth]{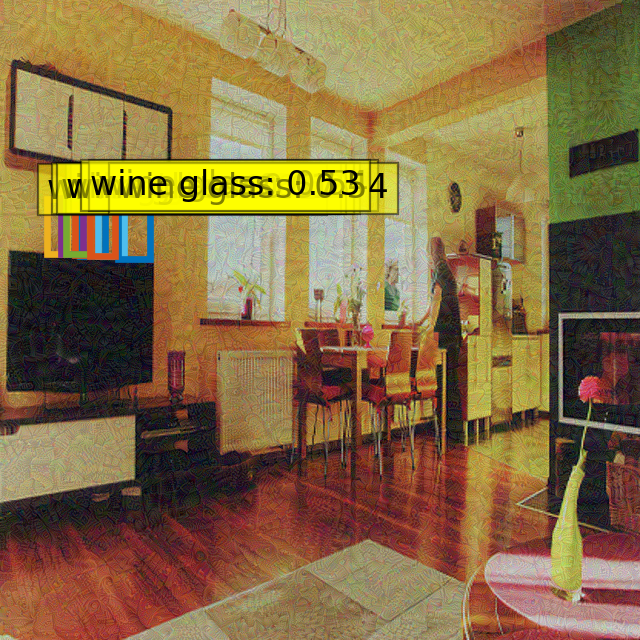}
        \caption{}
        \label{fig:example_focs_raw}
    \end{subfigure}
    \begin{subfigure}{0.45\linewidth}
        \centering
        \includegraphics[width=1.0\linewidth]{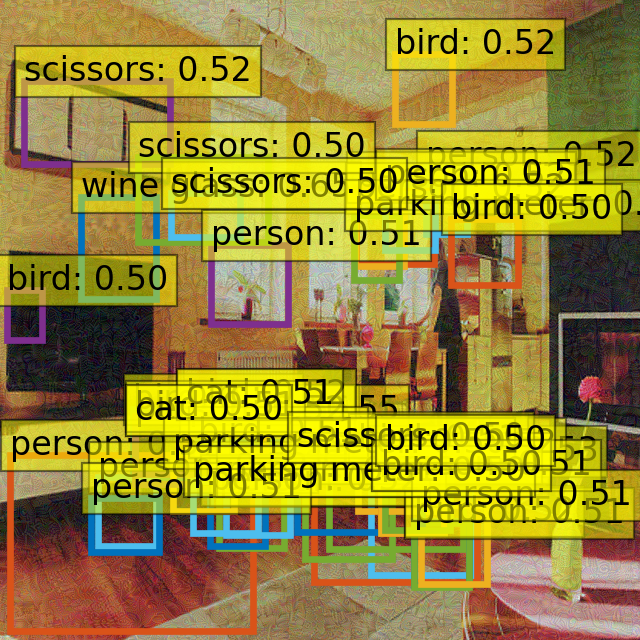}
        \caption{}
        \label{fig:example_focs_overload}
    \end{subfigure}
    \caption{The outputs of the adversarial examples by FCOS. \ref{fig:example_focs_raw} and \ref{fig:example_focs_overload} are generated by the normal PGD attack and Overload attack, respectively. }
    \label{fig:example_focs}
\end{figure}

We would like to emphasize that the proposed loss function defined in (4) is tailored specifically for the YOLO series. As detailed in the background section, each model employs its own unique algorithm to process the locations and probabilities of the output objects. For instance, two-stage detectors divide the task into two phases, resulting in the inability to directly estimate the gradient. Some detectors calibrate the locations based on predefined anchors. Nevertheless, there could be significant benefits in adjusting spatial importance.
Fig. \ref{fig:example_retinanet} and \ref{fig:example_focs} illustrate the outputs of adversarial examples generated by Retinanet \cite{lin2017focal} and FCOS \cite{tian2019fcos}, respectively. As can be seen, the predictions of adversarial examples produced by the standard PGD attack tend to cluster within a small region while the proposed attack ensures a more widespread distribution of objects in the spatial domain. We believe the spirit of Overload is applicable to most detectors integrated with NMS, but the implementation details should be further studied.

\begin{table*}[tb]
    \centering
    \begin{tabular}{ccccc|cccc}
        \toprule
        \multirow{3}{*}{Percentile} & \multicolumn{4}{c|}{YOLOv7} & \multicolumn{4}{|c}{YOLOv7-tiny} \\ 
        \cline{2-9}
        & \multicolumn{2}{c}{Adversarial Examples}  & \multicolumn{2}{c|}{Original Examples}
        & \multicolumn{2}{|c}{Adversarial Examples} & \multicolumn{2}{c}{Original Examples} \\
        & objects & boxes & objects & boxes & objects & boxes & objects & boxes \\
        \midrule
        min    & 18310 & 3139 &  0 & 0    &    18884 & 2398 &  0 & 0  \\
        0.10   & 19897 & 3765 &  6 & 1    &    23018 & 3259 & 97 & 9  \\
        0.25   & 20119 & 4069 & 18 & 2    &    23068 & 3390 & 32 & 3  \\
        0.50   & 20578 & 3681 & 30 & 3    &    23163 & 3313 & 12 & 1  \\
        0.75   & 21310 & 4991 & 10 & 2    &    23083 & 3519 & 28 & 3  \\
        0.90   & 20359 & 3912 & 27 & 3    &    23007 & 3708 & 74 & 9  \\
        max    & 22072 & 4243 & 12 & 1    &    22971 & 3399 & 36 & 2  \\
        \bottomrule
    \end{tabular}

    \begin{tabular}{ccccc|cccc}
        \toprule
        \multirow{3}{*}{Percentile} & \multicolumn{4}{c|}{YOLOv8s} & \multicolumn{4}{|c}{YOLOv8n} \\ 
        \cline{2-9}
        & \multicolumn{2}{c}{Adversarial Examples}  & \multicolumn{2}{c|}{Original Examples}
        & \multicolumn{2}{|c}{Adversarial Examples} & \multicolumn{2}{c}{Original Examples} \\
        & objects & boxes & objects & boxes & objects & boxes & objects & boxes \\
        \midrule
        min    & 8244 & 1188 &   0 &  0    &    7881 & 1652 &   0 &  0  \\
        0.10   & 8273 & 1273 &  46 &  6    &    8044 & 1747 &  56 &  9  \\
        0.25   & 8250 & 1098 &  28 &  4    &    7839 & 1530 &  10 &  1  \\
        0.50   & 8313 & 1212 &  76 & 14    &    7759 & 1884 &   7 &  1  \\
        0.75   & 8333 & 1326 & 133 & 21    &    7886 & 1918 &   1 &  1  \\
        0.90   & 8333 & 1141 &  51 &  5    &    8017 & 1658 & 121 & 23  \\
        max    & 8347 & 1268 &  37 &  5    &    7968 & 1836 &  85 & 16  \\
        \bottomrule
    \end{tabular}

    \begin{tabular}{ccccc|cccc}
        \toprule
        \multirow{3}{*}{Percentile} & \multicolumn{4}{c|}{YOLOv3} & \multicolumn{4}{|c}{YOLOv3-tiny} \\ 
        \cline{2-9}
        & \multicolumn{2}{c}{Adversarial Examples}  & \multicolumn{2}{c|}{Original Examples}
        & \multicolumn{2}{|c}{Adversarial Examples} & \multicolumn{2}{c}{Original Examples} \\
        & objects & boxes & objects & boxes & objects & boxes & objects & boxes \\
        \midrule
        min    & 11879 & 2107 &   0 &  0    &    5760 & 1554 &   0 &  0  \\
        0.10   & 13198 & 2558 &  76 &  7    &    5873 & 1674 &   0 &  0  \\
        0.25   & 13401 & 2565 &  62 &  6    &    5884 & 1619 &  56 &  6  \\
        0.50   & 14015 & 2562 & 134 &  9    &    5940 & 2061 &  30 &  3  \\
        0.75   & 14109 & 2707 & 205 & 17    &    5867 & 2380 & 125 & 24  \\
        0.90   & 14365 & 2839 &  84 &  7    &    5981 & 2528 &  16 &  4  \\
        max    & 14172 & 2651 &  39 &  3    &    5870 & 1543 &  21 &  2  \\
        \bottomrule
    \end{tabular}
    \caption{Latency Attacks on YOLOv7, YOLOv8, and YOLOv3 models}
    \label{table:nms_nx_yolo}

\end{table*}

\section{Transfer Attack Evaluation}

\begin{table*}[tb]
    \centering

    \begin{tabular}{ccccccc|cccc}
        \toprule
        \multirow{3}{*}{Percentile} & \multicolumn{6}{c|}{YOLOv5s} & \multicolumn{4}{|c}{YOLOv3} \\ 
        \cline{2-11}
        & \multicolumn{3}{c}{Adversarial Examples}  & \multicolumn{3}{c|}{Original Examples}
        & \multicolumn{2}{|c}{Adversarial Examples} & \multicolumn{2}{c }{Original Examples} \\
        & objects & boxes & time & objects & boxes & time & objects & boxes & objects & boxes \\
        \midrule
        min    & 10778 & 1541 &  11.1 &    0 &  0 & 14.1   &     8443 & 1107 &   0 &  0 \\
        0.10   & 15771 & 1897 &  43.8 &   40 &  3 & 16.4   &    13138 & 1518 &  27 &  4 \\
        0.25   & 18167 & 2124 &  96.2 &   47 &  4 & 16.4   &    14091 & 1563 & 220 & 18 \\
        0.50   & 22755 & 2392 & 132.7 &   28 &  3 & 16.4   &    14796 & 1629 &  49 &  5 \\
        0.75   & 20338 & 1844 & 170.4 &    3 &  1 & 16.6   &    15418 & 1661 &  44 &  3 \\
        0.90   & 22384 & 2795 & 248.3 &  234 & 21 & 16.7   &    16027 & 1656 &  63 &  3 \\
        max    & 23579 & 1580 & 252.9 &  212 & 21 & 16.9   &    17684 & 1785 &  12 &  1 \\
        \bottomrule
    \end{tabular}
    \caption{Results of the ensemble attack}
    \label{table:transfer}

\end{table*}

\begin{figure}[t]
    \centering
    \begin{subfigure}{.45\linewidth}
        \centering
        \includegraphics[width=1.0\linewidth]{fig_pgd_x_org.png}
        \caption{Original image \\ \ }
    \end{subfigure}
    \begin{subfigure}{0.45\linewidth}
        \centering
        \includegraphics[width=1.0\linewidth]{fig_pgd_x_adv.png}
        \caption{Adversarial image \\ \ }
    \end{subfigure}
    \begin{subfigure}{.45\linewidth}
        \centering
        \includegraphics[width=1.0\linewidth]{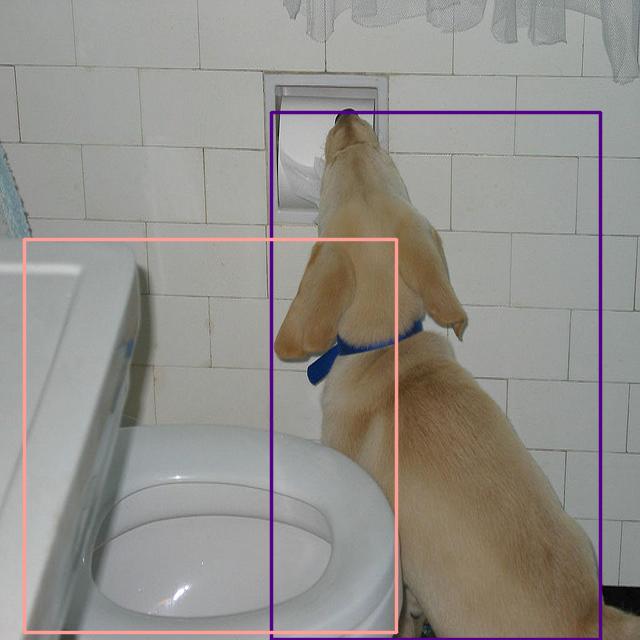}
        \caption{The output of the original image}
    \end{subfigure}
    \begin{subfigure}{0.45\linewidth}
        \centering
        \includegraphics[width=1.0\linewidth]{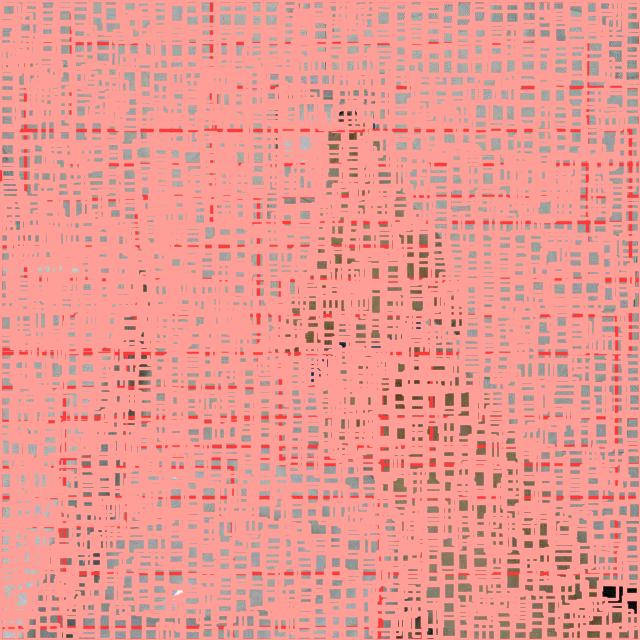}
        \caption{The output of the adversarial image}
    \end{subfigure}
    \caption{An example of Overload attack for object detection.}
    \label{fig:example}
\end{figure}

\begin{figure}[t]
    \centering
    \begin{subfigure}{.45\linewidth}
        \centering
        \includegraphics[width=.99\linewidth]{fig_ensemble_org.png}
        \caption{Original image \\ \ }
    \end{subfigure}
    \begin{subfigure}{0.45\linewidth}
        \centering
        \includegraphics[width=.99\linewidth]{fig_ensemble_adv.png}
        \caption{Adversarial image of the ensemble attack}
    \end{subfigure}
    \begin{subfigure}{.45\linewidth}
        \centering
        \includegraphics[width=.99\linewidth]{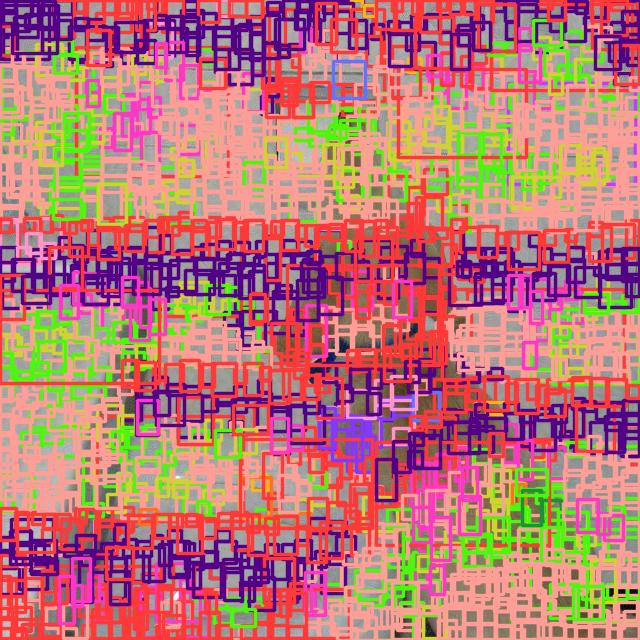}
        \caption{The output of YOLOv3 }
    \end{subfigure}
    \begin{subfigure}{0.45\linewidth}
        \centering
        \includegraphics[width=.99\linewidth]{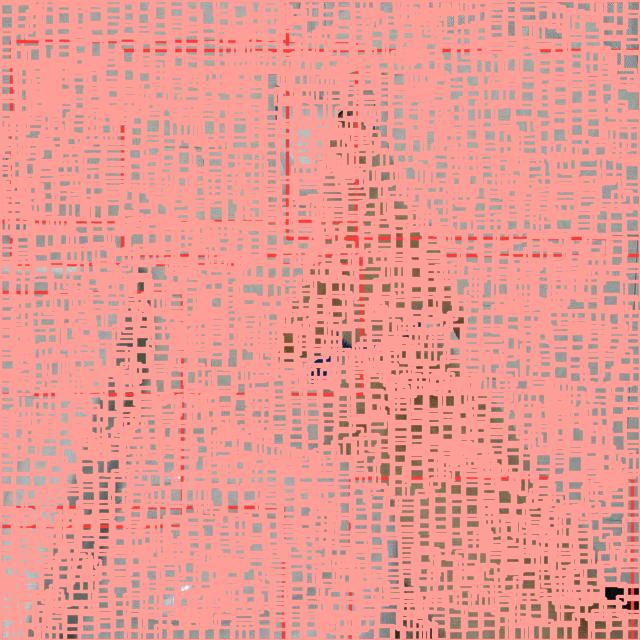}
        \caption{The output of YOLOv5s}
    \end{subfigure}
    \caption{An example of ensemble attack for object detection.}
    \label{fig:transfer_example}
\end{figure}

The transferability of adversarial attacks, where an attack crafted for one model can be successfully applied to a different victim model, is a common phenomenon in image classification tasks. However, when testing the transferability among different models in the YOLOv5 family for the latency attack, we did not observe this property. One possible reason is that the object detection network utilizes the Feature Pyramid Network (FPN), which combines features extracted from both low-resolution and high-resolution sources. This integration of features from different networks may lead to divergence and hinder the transferability of the attack.

Nevertheless, we explored an alternative approach known as ensemble training to craft adversarial examples that can deceive multiple models. In the ensemble attack, gradients are obtained from either one candidate model or averaged across all candidate models in each attack step. We evaluated the performance of the ensemble attack using a combination of YOLOv3 and YOLOv5s models. Table \ref{table:transfer} presents the results of the ensemble attack, omitting the execution times for YOLOv3 due to a technical issue with compiling the model to TensorRT format.

As observed, the ensemble attack successfully generates a significant number of objects for both YOLOv3 and YOLOv5s simultaneously. However, comparing these results with those in Table \ref{table:nms_nx}, it appears that the strength of the ensemble attack is slightly weaker than that of the native attack. To provide visual context, Figure \ref{fig:example} and Figure \ref{fig:transfer_example} illustrate the original image, the corresponding adversarial image, and the results obtained from Overload and the ensemble attack, respectively.

These findings suggest that information from multiple models can be encoded within a single image, enabling the ensemble attack to deceive different object detection models. However, further investigation is needed to enhance the effectiveness and transferability of the ensemble attack against the latency-based defense.
\end{document}